\documentclass[runningheads]{llncs}
\usepackage{graphicx}
\usepackage{amsmath,amssymb} 
\usepackage{color}

\usepackage{hyperref}
\usepackage{epsfig}
\usepackage{ifthen}

\usepackage{float}
\usepackage[caption = false]{subfig}
\newcommand*{\affmark}[1][*]{\textsuperscript{#1}}

\definecolor{gray}{rgb}{0.5,0.5,0.5}
\definecolor{green}{rgb}{0, 0.4, 0}
\definecolor{orange}{rgb}{1, 0.4, 0}
\definecolor{mahogany}{rgb}{0.75, 0.25, 0.0}
\definecolor{purple}{rgb}{0.6, 0, 0.6}
\definecolor{darkgreen}{rgb}{0, 0.4, 0}
\definecolor{frenchblue}{rgb}{0.0, 0.45, 0.73}
\definecolor{crimson}{rgb}{0.6, 0.0, 0.0}
\newboolean{revising}
\setboolean{revising}{False}
\ifthenelse{\boolean{revising}}
{
	\newcommand{\ignore}[1]{}

    \newcommand{\gina}[1]{{\color{purple}{#1}}}
	\newcommand{\ginarep}[2]{{\color{purple}{#2}}}

    \newcommand{\jc}[1]{{\color{frenchblue}{#1}}}
	
}
{
	\newcommand{\ignore}[1]{}

    \newcommand{\gina}[1]{#1}
	\newcommand{\ginarep}[2]{#2}

    \newcommand{\jc}[1]{#1}

}

\begin{document}
\pagestyle{headings}
\mainmatter

\title{Liquid Pouring Monitoring via \\Rich Sensory Inputs}

\titlerunning{Liquid Pouring Monitoring via Rich Sensory Inputs}

\authorrunning{T.Y. Wu\affmark[$\star$], J.T. Lin\affmark[$\star$], T.H. Wang, C.W. Hu, J.C. Niebles, M. Sun}

\author{Tz-Ying Wu\inst{1}\affmark[,$\star$]
\and Juan-Ting Lin\inst{1}\affmark[,]\thanks{indicates equal contribution}
\and Tsun-Hsuang Wang\inst{1}
\and Chan-Wei Hu\inst{1}
\and \\Juan Carlos Niebles\inst{2} 
\and Min Sun\inst{1}
}

\institute{
Department of Electrical Engineering, National Tsing Hua University, Taiwan\\
	\email{ \{gina9726, brade31919, johnsonwang0810, huchanwei1204\}@gmail.com, sunmin@ee.nthu.edu.tw}
\and Department of Computer Science, Stanford University, USA\\
    \email{jniebles@cs.stanford.edu}
}

\maketitle

\begin{abstract}
Humans have the amazing ability to perform very subtle manipulation task using a closed-loop control system with imprecise mechanics (i.e., our body parts) but rich sensory information (e.g., vision, tactile, etc.). In the closed-loop system, the ability to monitor the state of the task via rich sensory information is important but often less studied.
In this work, we take liquid pouring as a concrete example and aim at learning to continuously monitor whether liquid pouring is successful (e.g., no spilling) or not via rich sensory inputs.
We mimic humans' rich sensories using synchronized observation from a chest-mounted camera and a wrist-mounted IMU sensor.
Given many success and failure demonstrations of liquid pouring, we train a hierarchical LSTM with late fusion for monitoring.
To improve the robustness of the system, we propose two auxiliary tasks during training: inferring (1) the initial state of containers and (2) forecasting the one-step future 3D trajectory of the hand with an adversarial training procedure.
These tasks encourage our method to learn representation sensitive to container states and how objects are manipulated in 3D. 
With these novel components, our method achieves $\sim 8\%$ and $\sim 11\%$ better monitoring accuracy than the baseline method without auxiliary tasks on unseen containers and unseen users respectively.

\end{abstract}

\section{Introduction}
\label{sec:intro}

\gina{
Researchers in cognitive science community have conducted several studies~\cite{probsim,probsim2} of mental simulation, and proved that humans have some internal mechanisms to reason daily life physics with relative ease.}
\gina{Some robotics research borrows a hand from human demonstrations to tackle manipulation problems; for example, recently, Edmonds et al.~\cite{feeling} leverage multimodal sensor to capture poses and contact forces to learn the manipulation of opening medicine bottles.
Humans can be viewed as closed-loop control systems with imprecise mechanics (i.e., our body parts) but rich sensory information (e.g., vision, tactile, etc.). The sensory feedback helps us continuously reason the environment, and plan our next action according to it.
In the closed-loop system, the ability to monitor the state of the task via rich sensory information is important but often less studied. Monitoring subtle manipulation task is useful for both in-home elder care system and virtual training in medical scenarios (e.g., training surgical operation), since a system with this kind of ability can further assist people to accomplish subtle tasks.}

\gina{Liquid pouring is a subtle manipulation task that humans learn during childhood and can easily perform on a daily basis. This task requires continuously monitoring environmental states such as the liquid level in containers and the relative position and motion between containers in order to adjust future actions toward not spilling. For instance, if the receiver container is empty and the source container is tilting slowly, one should speed-up the tilting action. In contrast, if the receiver container is almost full and the source container is tilting fast, one should slow down the tilting action to prevent overflow. This suggests that both object states, relative position and motion are very important cues for subtle manipulation tasks such as liquid pouring.
With the ability to monitor liquid pouring, an intelligent system can either stop the user from spilling, or bring a duster to the user when the liquid is spilled.

Monitoring liquid pouring activity is a very subtle task compared to mainstream activity recognition tasks such as action classification or temporal detection~\cite{Youtube-8M,ActivityNet}. Hence, only a few works have made progress toward this direction in computer vision. 
Alayrac et al.~\cite{state-estimation-3} propose to discover object states and manipulation actions in videos. However, they only consider empty versus full (binary) container states and multiple discrete actions where pouring is one of them.
Recently, Mottaghi et al.~\cite{Pouring} propose to reason about volume and content in liquid containers to predict how much liquid will remain in the container if we tilt it by $x$ degrees (referred to as pouring prediction). However, we argue that such prediction target has limited application since it does not directly answer how to pour liquid successfully or whether the pouring action results in success or failure. 

In this work, we take liquid pouring as a concrete example and aim at learning to continuously monitor whether liquid pouring is successful (e.g., not spilling) or not via rich sensory inputs.
Cognitive scientists suggest that people have the ability to simulate pouring behaviors in their mind, which is mentioned in \cite{probsim}. However, there remain discrepancies between the simulation and the real results. By continuously observe current environmental states, people can adjust their ways to manipulate the object (e.g. the angle of the container) in order to reach their goal. This process can be viewed as a closed-loop control.
In order to borrow a hand from humans' physical reasoning ability, we mimic humans' rich sensors using synchronized observation from a chest-mounted camera and a wrist-mounted IMU sensor as the input (details in \autoref{sec:dataset}). The target output for monitoring is a binary class: a success or a failure pouring trial.} 
\gina{To study liquid pouring monitoring in the real world by leveraging human demonstrations, we collect a liquid pouring dataset containing both successful and failed demonstrations with all inputs and outputs information mentioned above. 
To the best of our knowledge, this is the first dataset with multimodal sensor information for studying monitoring in a subtle liquid pouring task.
}

\begin{figure}[t!]
\begin{center}
\includegraphics[width=0.7\linewidth]{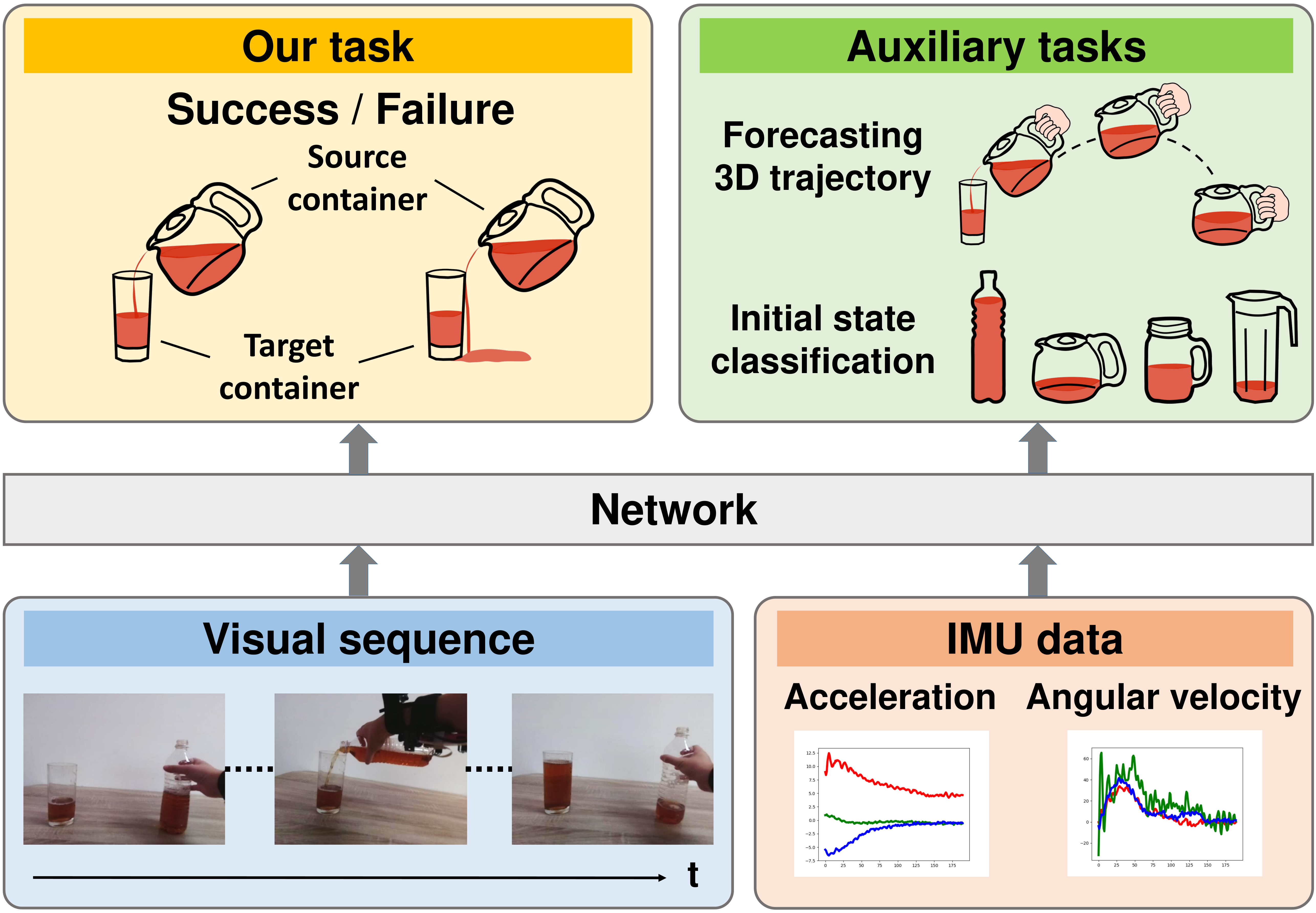}
\caption{\textbf{Overview.} From a series of visual observations and IMU data, our model can monitor if this sequence is a success or failure with two auxiliary tasks: initial object state classification (different containers with different initial liquid levels) to ensure the recurrent model encode states sensitive features; forecasting 3D trajectory requires the ability to model hand dynamics during the pouring process, providing a strong cue for our monitoring task.  The details of auxiliary tasks are described in \autoref{sec:aux}}
\label{fig:overview}
\end{center}
\end{figure}

\gina{Given many success and failure demonstrations of liquid pouring,
we train a hierarchical LSTM~\cite{MRNN} with late fusion to incorporate rich sensories inputs without significantly increasing the model parameters as compared to early fusion models.
To further improve the generalizability of our method, we introduce two auxiliary tasks during training: (1) predicting the initial state of containers and (2) forecasting the one-step future 3D trajectory of the hand with an adversarial training procedure.
These auxiliary tasks encourage our method to learn representation sensitive to container states and how objects are manipulated in 3D. 
In our experiments, our method achieves $\sim 8\%$ and $\sim 11\%$ better monitoring accuracy than the baseline method without auxiliary tasks on unseen containers and unseen users respectively.
}

\section{Related Work}
\label{sec:relatedwork}

\noindent\textbf{Activity Recognition.}
Activity recognition has received lots of attention from the computer vision community and already has many released datasets \cite{UCF101,HMDB,Youtube-8M,ActivityNet,AVA} containing diverse actions. Many prior works on activity recognition focus on understanding human activity through observing body poses {\cite{body-pose-1,P-CNN,body-pose-2}}, scenes {\cite{scene-1,scene-2}} or objects interacting with human {\cite{interact-1,interact-2,interact-3,interact-4}}. There are also many works {\cite{RW-other-works-1,RW-other-works-2,RW-other-works-3}} considering recognizing activity through egocentric videos, some of which use depth sensor \cite{activityrecognition_RGBD,fine-grained-RGBD} as well in attempt to enhance the perception of the changes in the environment. There are also methods \cite{song2016egocentric} and datasets \cite{Torre2009CMU-MMAC,opportunity} utilizing multimodal sensor inputs to perform activity recognition. These established datasets mainly focus on diverse activity recognition and  do not include failure cases. However, we focus more on distinguishing subtle differences among behaviors targeting on the same objective (liquid pouring). Therefore, we collect our own liquid pouring dataset with multimodal sensor data which includes both success and failure cases (details in \autoref{sec:dataset}).  

\noindent\textbf{Fine-grained activity recognition.}
Many methods focused on interacting and manipulating motions between human and objects.  Lei \textit{et al.}\cite{fine-grained-RGBD} applied RGB-D camera to achieve the robust object and action recognition. There are also methods utilizing spatiotemporal information \cite{fine-grained-midlevel,fine-grained-pipeline,fine-grained-fisher,spatio-temperal-2018,spatio-temperal-2018-2}. By combining spatiotemporal and object semantic features, Yang \textit{et al.}\cite{fine-grained-midlevel} find key interaction without using further object annotations. 
In this work, rather than designing special procedures to mine unique spatiotemporal features, we introduce auxiliary tasks to learn feature good for multiple tasks.  

\noindent\textbf{Environmental State Estimation.}
In liquid pouring sequences, container and the liquid state can be estimated from RGB inputs. Alayrac \textit{et al.}\cite{state-estimation-3} model the interaction between actions and objects in a discrete manner. Some methods further demonstrate that liquid amount can be estimated by combining semantic segmentation CNN and LSTM \cite{state-estimation-1,Pouring}. In contrast, our main goal is not to explicitly recognize environmental states. We aim at implicitly learning environmental state sensitive features such that our performance in monitoring can be improved. Recently, Sermanet \textit{et al.}~\cite{SermanetLHL17} also propose to learn states sensitive feature in a self-supervised manner.  

\noindent\textbf{Robot Liquid Pouring.}
In the robotics community, there are a number of works~\cite{stereo,TAMOSIUNAITE2011910,Rozo,BrandiK014,SchenckF16c,pouring-5,pouring-6} directly tackle the manipulating task of liquid pouring without considering the monitoring task.\gina{~\cite{stereo} build a liquid dynamic model using optical flow.}~\cite{pouring-5,pouring-6} are developed in synthetic environments.
Tamosiunaite \textit{et al.}~\cite{TAMOSIUNAITE2011910} apply model-based reinforcement learning. Rozo \textit{et al.}~\cite{Rozo} propose a parametric hidden Markov model to direct regress control commands. Brandl \textit{et al.}~\cite{BrandiK014} learn to generalize pouring to unseen containers by warping the functional parts of the unseen containers to mimic the functional parts of a seen container. 
Schenck and Fox~\cite{SchenckF16c} propose to first estimate the volume of liquid in a container; then, a simple PID controller is used to pour specific amounts of liquid. 
However, all of the methods above are not evaluated on generalization jointly across users, containers states, container instances.

\section{Overview}
\label{sec:method}
In this section, we first formulate the problem of monitoring liquid pouring. Next, we describe our recurrent model for fusing multimodal data. Our method with two auxiliary tasks will be mainly described in \autoref{sec:aux}.

\begin{figure*}[t]
\centering
\includegraphics[width=0.95\textwidth]{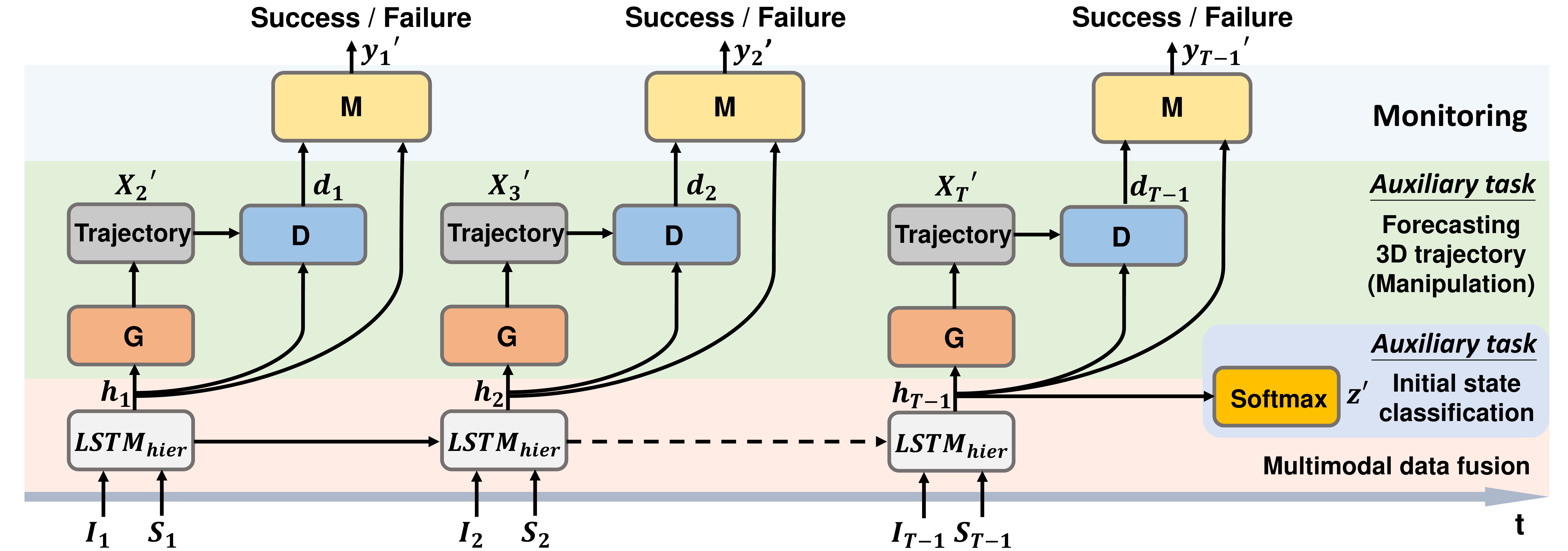}
\caption{\textbf{Model architecture.} Our model consists of a hierarchical LSTM $LSTM_{hier}$ (details in \autoref{ssec:LSTM}), a generator $G$, a discriminator $D$ and a monitoring module $M$ (details in \autoref{sec:aux}). There are two auxiliary tasks in our method, which are 3D trajectory forecasting (green shading) and initial state classification (blue shading). At each time step $t$, $LSTM_{hier}$ will encode visual observation $I_t$ and IMU data $S_t$ to $h_t$ (red shading). $G$ will generate a trajectory $X_{t+1}^{\prime}$ according to hidden encoding $h_t$. $D$ will distinguish if the input trajectory is generated or not corresponding to $h_t$, which models the dynamics during the manipulation. $M$ will predict if this pouring sequence is a success or failure based on the discriminator score $d_{t}$ and hidden encoding $h_{t}$. At the end of the sequence, the model will classify $36$ initial states as an auxiliary task
}\label{fig:model}
\end{figure*}

\subsection{Problem Formulation}
\noindent\textbf{Notations.}
For all of our notations, general font style stands for ground truth data, and prime stands for predictions. For example, $y_t$ is the ground truth label for whether the sequence is a success and $y_t^{\prime}$ is the prediction. Notations with boldface denote a sequence of data. $t$ denotes a certain time step, and $T$ stands for the total time steps of the sequence.
\\
\noindent\textbf{Observation.}
To capture visual and motion information like liquid content, container type and dynamics of the demonstrator's hand during the pouring process, we use a multimodal sensing system including a camera on the front chest and an IMU sensor on the wrist. At each time step $t$, the camera observes visual observation $I_t$, and the 6DOF IMU sensor captures motion observation $S_{t}=\{\mathbf{a}^{1}, \mathbf{a}^{2},...,\mathbf{a}^{N}\}$, where $\mathbf{a}^{i}$ is the $i$'th sample in the current time step, $i\in1\sim N$, and $N$ denotes the number of samples in this time step. In practice, $N=38$, i.e., IMU sensor will capture $38$ samples within two consecutively captured camera frames. $\mathbf{a}=\{a_{1}, a_{2}, a_{3}, a_{4}, a_{5}, a_{6}\}$ is a single piece of real-valued data from the IMU, where $(a_{1}, a_{2}, a_{3})$ is the acceleration and $(a_{4}, a_{5}, a_{6})$ is the angular velocity corresponding to $x$, $y$, and $z$ axis. Simultaneously, at each time step $t$, we obtain hand 3D trajectory ground truth $X_{t}=(P, R)$ by a HTC Vive tracker mounted on the wrist, where $P=(p_{x}, p_{y}, p_{z})$ and $R=(r_{x}, r_{y}, r_{z})$ stand for the position part and rotation part in world coordinate respectively. Note that HTC Vive system is only used in training.
\\
\noindent\textbf{Goal.}
In our task, we aim at learning to monitor whether the pouring liquid sequence is a success or failure with two auxiliary tasks, which are initial object state classification (IOSC) and next-step hand 3D trajectory forecasting (TF). Considering the input sequence containing visual images $\mathbf{I}=\{I_{1}, I_{2},...,I_{T}\}$ and IMU data $\mathbf{S}=\{S_{1}, S_{2},...,S_{T}\}$, the output of our model for each time step $t$ are the prediction $y_{t}^{\prime}$ indicating whether the sequence is a success for our monitoring task and next-step trajectory prediction $X_{t+1}^{\prime}$ for 3D trajectory forecasting, where $t\in1\sim T-1$, $T$ denotes the total time steps of the sequence. In the end of the whole sequence, our model will predict the initial object state $z^\prime$ of the sequence among the $36$ variations (details in \autoref{sec:dataset}).

\begin{figure}[t!]
\begin{minipage}[c]{0.45\textwidth}
\includegraphics[width=\linewidth]{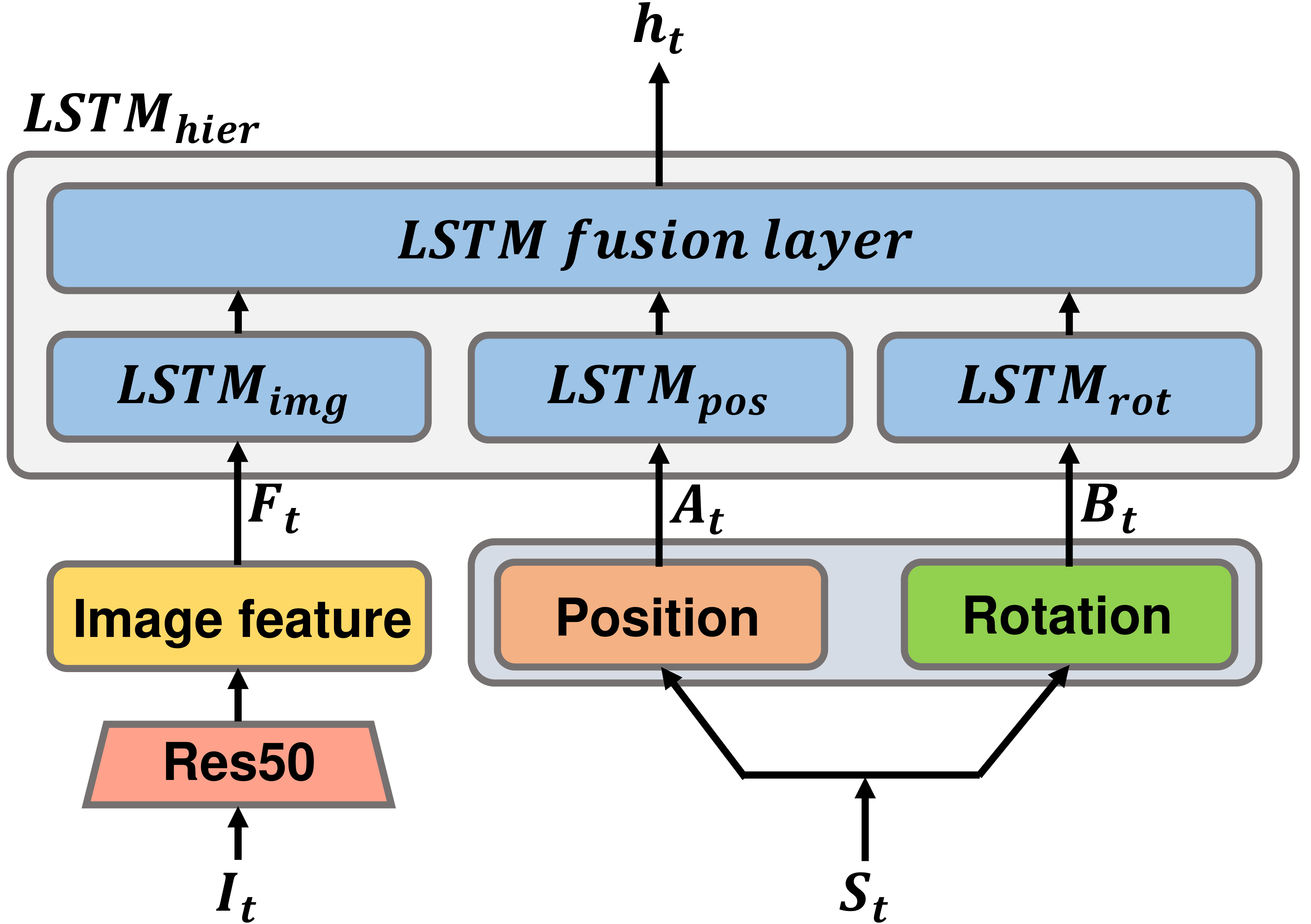}
\end{minipage}
\begin{minipage}[c]{0.05\textwidth}
\end{minipage}
\begin{minipage}[c]{0.5\textwidth}
\caption{\textbf{LSTM encoder.} Our hierarchical LSTM encoder $LSTM_{hier}$ consists of 3 LSTM cells ($LSTM_{img}$, $LSTM_{pos}$, $LSTM_{rot}$) at the first level and a LSTM fusion layer to fuse these hidden encodings at the second level, fusing multimodal inputs containing image feature $F_{t}$, hand position feature $A_{t}$ and hand rotation feature $B_{t}$ computed from IMU sensor}
\label{fig:lstm}
\end{minipage}
\end{figure}

\subsection{Multimodal Data Fusion}
\label{ssec:LSTM}
To catch and combine the temporal sequence of input from image and IMU sensor, we adopt a hierarchical LSTM \gina{proposed by \cite{MRNN}} \ginarep{$LSTM_{hier}$ (see \autoref{fig:lstm})}{}to handle scale differences among multimodal inputs. In the first layer of our module $LSTM_{hier}$ (see \autoref{fig:lstm}), there are 3 LSTM cells ($LSTM_{img}$, $LSTM_{pos}$, $LSTM_{rot}$) with different hidden layer sizes to encode the inputs from three different sources:
(1) image feature $F_{t}=\mathbf{Res50}(I_{t})$ extracted from the pool5 layer of ResNet50 \cite{ResNet50} with dimension of $1\times 2048$,
(2) hand position feature: the aggregation of acceleration along 3 axis 
$A_{t}={\{(a_{1}^{i}, a_{2}^{i}, a_{3}^{i})\}_{i=1}^{N}}\subset S_{t}$ with dimension of $1\times 3N$ and
(3) hand rotation feature: the aggregation of angular velocity along 3 axis $B_{t}={\{(a_{4}^{i}, a_{5}^{i}, a_{6}^{i})\}_{i=1}^{N}}\subset S_{t}$ with dimension of $1\times 3N$.
Then the encoded features are concatenated as the input to the second layer consisting of a single LSTM cell. The output encoded feature $h_{t}=LSTM_{hier}(F_{t},A_{t},B_{t})$ of the hierarchical LSTM will be passed to the generator $G$, discriminator $D$ and the monitor module (please refer to \autoref{sec:aux}). 

\section{Monitoring with Auxiliary Tasks}
\label{sec:aux}

Monitoring the success of a pouring sequence is a challenging task since subtle changes in states of the environment are hard to perceive. Intuitively, the initial object state and the hand dynamics are the strong cues for monitoring pouring process. We model the object and manipulator (i.e., hand) states implicitly by a hierarchical LSTM $LSTM_{hier}$ and introduce two auxiliary tasks, 3D trajectory forecasting (TF) and initial object state classification (IOSC).
In this section, we describe the details of the two auxiliary tasks and our monitoring module. 

\subsection{Forecasting 3D Trajectory}
\label{ssec:tf}
Forecasting 3D trajectory is a path for us to learn to model the dynamics of the manipulator during the pouring sequence. The most naive way to predict trajectory is to train direct regression on demonstration sequences; however, the generated trajectory will be very limited to the data distribution of training data as the amount and the diversity of training data is limited. To model the distribution of successful demonstration and to generate more diverse trajectories, we introduce adversarial training loss $L_{adv}$ proposed by Goodfellow \textit{et al.} \cite{GAN} here with a generator $G$ to generate trajectory prediction and a discriminator $D$ to distinguish if the input trajectory is generated or not (see \autoref{fig:model}).

\noindent\textbf{Generator.}
Taking the encoded feature $h_{t}$ from $LSTM_{hier}$ as input, our generator predicts next-step trajectory $X_{t+1}^{\prime}=G_{\theta_{G}}(h_t)$ as output, where $G_{\theta_{G}}$ is a three-layer fully-connected feed-forward network parametrized by $\theta_{G}$.
Our generator has two objectives: 

(1) Generate the trajectory which is close to the ground truth demonstration. (modeled by the regression loss).
(2) Fool discriminator with the generated trajectory (modeled by the adversarial loss).

Thus, our loss function for the generator can be derived as follows,
\begin{align}
\label{eq:gloss}
L_{Gen}=L_{reg}+\lambda * L_{adv},
\end{align}
where $\lambda$ is the weighting between the two different losses (we empirically set $\lambda$ to 1), $L_{reg}$ is the regression loss, and $L_{adv}$ stands for the adversarial loss.  

The regression loss is defined as follows,
\begin{equation}
\label{eq:regloss}
\begin{aligned}
L_{reg}=\frac{1}{T-1}\sum_{t=1}^{T-1}dist(X_{t+1},G_{\theta_{G}}(h_{t})), \\
\end{aligned}
\end{equation}
where $dist()$ is the distance function, $X_{t+1}$ is the ground truth trajectory, $G_{\theta_{G}}(h_{t})$ is the generated trajectory, and $T$ denotes the total time steps of the sequence. Recall the trajectory $X_{t+1}$ is composed of two parts, position $P=(p_{x}, p_{y}, p_{z})$ and rotation $R=(r_{x}, r_{y}, r_{z})$; likewise $G_{\theta_{G}}(h_{t})=(P^{\prime},R^{\prime})$, where $P^{\prime}=(p_{x}^{\prime}, p_{y}^{\prime}, p_{z}^{\prime})$, $R^{\prime}=(r_{x}^{\prime}, r_{y}^{\prime}, r_{z}^{\prime})$.  

The distance function is defined as
\begin{equation}
\begin{aligned}\label{eq:dist}
dist(X_{t+1},G_{\theta_{G}}(h_{t}))&=MSE(P,P^{\prime})+\sum_{k=x,y,z}(1-\cos{(r_k-r_k^{\prime})}),
\end{aligned}
\end{equation}
where MSE denotes Mean Squared Error. Here we use different distance metrics for rotation and translation because adopting cosine distance in angular difference is more reasonable. \gina{In particular, the cosine distance between $359^{\circ}$ and $0^{\circ}$ is small, but its mean square error is large. Note that we empirically adopt the same weighting for the position loss and rotation loss since the effect of different weightings is marginal on the performance.
}  

The adversarial loss is defined as follows,
\begin{equation}
\label{eq:adversarial}
L_{adv}=\frac{1}{T-1}\sum_{t=1}^{T-1}-\log{D_{\theta_{D}}(h_t,G_{\theta_{G}}(h_{t}))},
\end{equation}
where $D_{\theta_{D}}$ is the discriminator of our model and will be elaborated later.

\noindent\textbf{Discriminator.}
In training time, the discriminator takes both the encoded feature at that time step $h_t$ and the predicted trajectory $X_{t+1}^{\prime}=G_{\theta_{G}}(h_{t})$ from the generator or ground truth trajectory $X_{t+1}$ as inputs with the objective of catching generated trajectory from the generator. Adopting similar design from the generator, our discriminator $D_{\theta_{D}}$ is also modeled with a three-layer fully-connected feed-forward network parameterized by $\theta_{D}$. The discriminator loss is defined as follows, 
\begin{equation}
\begin{aligned}\label{eq:Dis}
L_{Dis}&=\frac{1}{T-1}\sum_{t=1}^{T-1}[-\log{(D_{\theta_{D}}(h_t,X_{t+1}))}-\log({1-D_{\theta_{D}}(h_t,G_{\theta_{G}}(h_{t}))})]
\end{aligned}
\end{equation}
In testing time, given the encoded feature $h_{t}$ and generated trajectory $X_{t+1}^{\prime}$ of the certain time step $t$, the discriminator will predict the score $d_{t}=D_{\theta_{D}}(h_{t}, X^{\prime}_{t+1}), t\in1 \sim T-1$ of whether the input sequence is generated or not.

\subsection{Initial Object State Classification}
\label{ssec:isc}
As we mention above, hand motion and initial object states are the two strong cues for monitoring pouring sequences. Learning the embedding of the data sequence is critical since the amount of training data is limited. To learn a good representation for  monitoring, we train the classification on the initial object state based on the hidden encoding from the hierarchical LSTM $LSTM_{hier}$ in the end of each successful demonstration sequence (see \autoref{fig:model}) as follows,
\begin{align}
\label{eq:cls}
&q=\textrm{Softmax}(\theta_{q}, h_{T-1}), \\
&z^{\prime}=\arg\max_{c\in\mathcal{Z}} q(c), \\
&L_{cls}=-\log{q(z)},
\end{align}
where $h_{T-1}$ is the hidden encoding at the last time step of the sequence, $\theta_q$ is the parameter of the classifier and $q\in R^{|\mathcal{Z}|}$ is the softmax probability of initial object states in $\mathcal{Z}$. $z^{\prime}$ is the prediction of the initial object state and $z$ denotes the ground truth initial object state. In our case, $|\mathcal{Z}|=36$, which means there are $36$ variations of initial object states (details can be referred to \autoref{sec:dataset}).

\subsection{Monitoring Module}
We propose a monitoring module $\textrm{M}$, which is designed as a single-layer network to predict whether a pouring sequence is a success or not given the hidden representation $h_{t}$ from $LSTM_{hier}$ and the discriminator score $d_{t}$ as inputs (see \autoref{fig:model}). The output of the monitoring module is defined as,
\begin{align}
\label{eq:mcls}
&y_{t}^{\prime}=\textrm{M}_{\theta_{M}}(h_{t}, d_{t}),
\end{align}
where $\theta_{M}$ is the parameter of $\textrm{M}$ and $y_{t}^{\prime}$ is the prediction of success or failure. We train our monitoring module with cross-entropy loss. The architecture of our monitoring module is compact and effective since our model has already learned powerful feature that can capture the appearance changes and hand dynamics during the pouring process through auxiliary tasks.  

\subsection{Implementation Details}
We use ResNet50\cite{ResNet50} trained on ImageNet\cite{imagenet} as the visual feature extractor. The input size of $LSTM_{img}$ is $2048$, and the input size of both $LSTM_{pos}$ and $LSTM_{rot}$ are $3N$ ($N=38$ in our case). $LSTM_{img}$ hidden size is $512$, and both $LSTM_{pos}$ and $LSTM_{rot}$ hidden size are $128$. The second layer of hierarchical LSTM has its hidden size $512$. Generator $G$ and discriminator $D$ are the 3-layered fully-connected network with each layer of size $128$. Monitor module is a fully-connected layer of size $256$. We train our model for $3000$ epochs with batch size $24$. Learning rate is $1e^{-4}$. We optimize all objectives with equal weightings.

\section{Dataset}\label{sec:dataset}

\begin{figure}[tb!]
\centering
\subfloat[Our multimodal sensing system]{\includegraphics[width=0.66\textwidth]{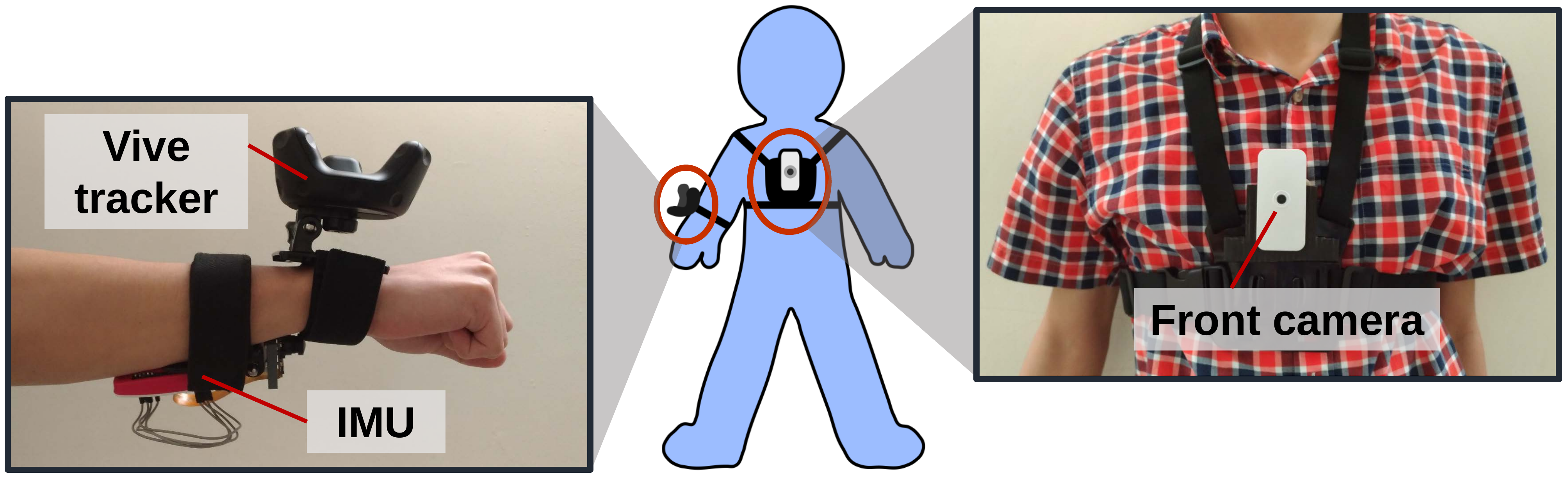}} \\
\subfloat[Variations of initial settings]{\includegraphics[width=0.66\textwidth]{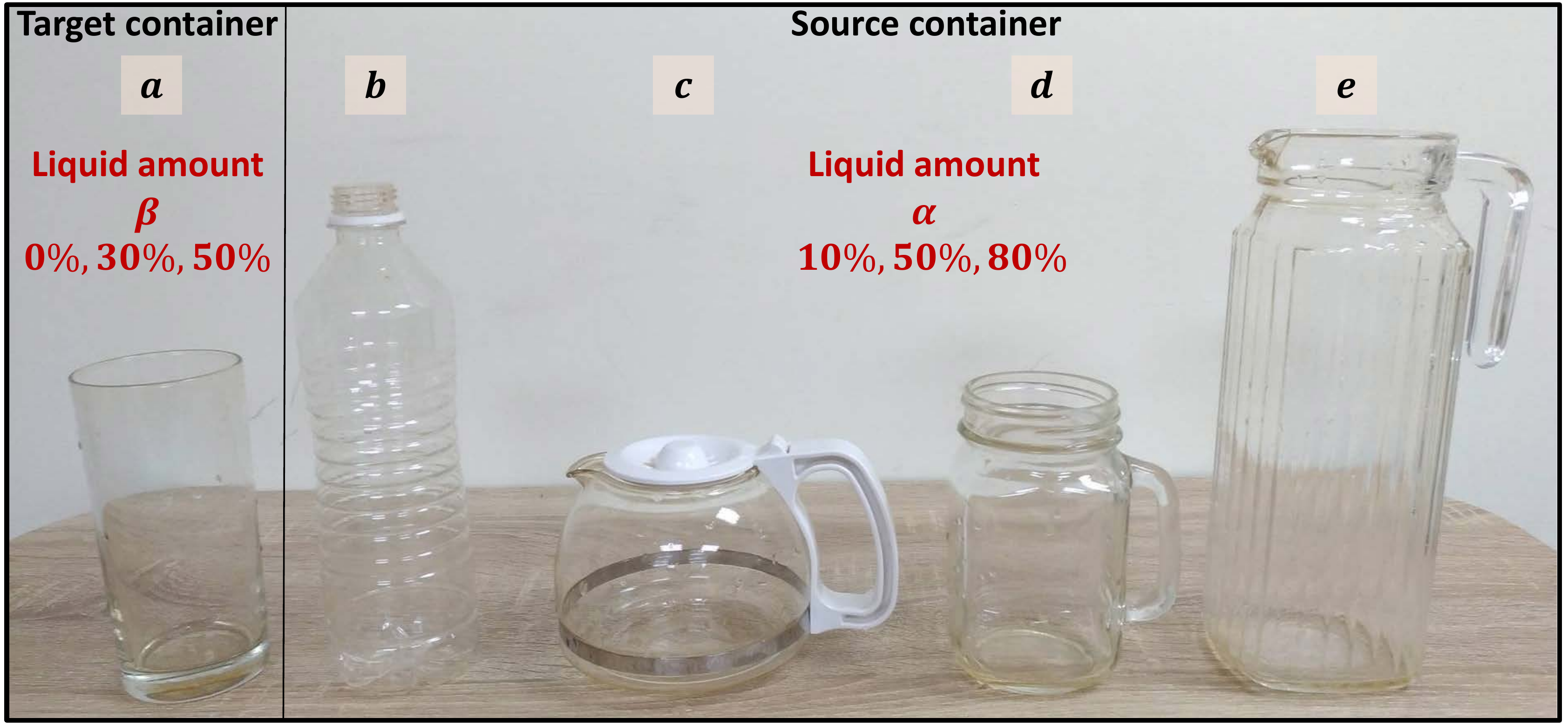}}
\caption{\textbf{Settings to collect our dataset.} (a) A camera is mounted on the chest to capture visual images. On the wrist, there is a vive tracker and an IMU sensor. (b) We use these containers to create variations of initial settings (details in \autoref{sec:dataset})}
\label{fig:setting}
\end{figure}

In order to examine our method on monitoring whether the pouring sequence belongs to successful / failure sequences, we collect both successful and failure pouring sequences with our multimodal sensing system. We have one chest-mounted camera to capture the first-person view observation; one wrist-mounted 6DOF IMU sensor and one tracker of the HTC Vive motion tracking system on the right wrist to catch both the motion observation and the ground truth trajectory simultaneously. \autoref{fig:setting}.a is the illustration of the devices on the demonstrator. \ginarep{In the following section, we will illustrate more on how we collect different kinds of demonstrations.}{We illustrate how we collect different kinds of demonstrations below.}
\\
\noindent \textbf{Variations of pouring sequences.}
Our single pouring sequence consists of pouring liquid from the source container with initial liquid amount $\alpha$ to target container with $\beta$ amount of liquid. \gina{Similar to~\cite{Pouring}, we roughly divide the container states into discrete labels.} In successful sequences, the demonstrator tries to fill target container with the liquid in the source container without spilling out any liquid. If target container is filled to about 80\% full, the demonstration will stop even if there is still liquid left in the source container. For single demonstrator, we will record the demonstrations with different kinds of containers and different initial liquid amounts to obtain more diverse demonstrations. For source container, we use 4 different containers $b, c, d, e$ in \autoref{fig:setting}.b with three different initial liquid amount $\alpha$: \{10\%, 50\%, 80\%\}. We use container $a$ in \autoref{fig:setting}.b as the target container with three different initial liquid amount $\beta$: \{0\%, 30\%, 50\%\}. Combining the different settings in source container, $\alpha$, and $\beta$, we can obtain total 36 different initial object states. In practice, we will record 5 repeated sequences for each initial object state setting. As a result, for a single demonstrator, we can obtain 180 demonstration sequences.

\noindent \textbf{Pouring \jc{styles}.}
In addition to different variations in the liquid amount and container appearances, we collect demonstrations conducted by 5 different demonstrators to ensure the diversity in pouring \jc{styles} from person to person.
\\
\noindent \textbf{Failure sequences.} 
In general, there can be many ways to conduct a failure sequence. However, to model the monitoring tasks, we choose one of the most common mistakes made by humans during the pouring sequences: \textit{Spill out} (The demonstrator accidentally spill out some liquid during the pouring action.)
Regarding the variations and pouring \jc{styles}, we use the same settings from the successful sequences: 
(1) 5 repeated sequences for each of 36 variations.
(2) 5 different demonstrators to ensure diverse pouring behaviors.  
\gina{Hence, the total amount of demonstration is $2*5*5*36=1800$.}

\section{Experiments}
\label{sec:experiment}

\gina{In this section, we introduce the evaluation metrics and settings used in our experiments. We then describe our monitoring experiments and discuss our experimental results with ablation studies.}

\subsection{Metrics}\label{ssec:metric}
In our experiments, we observe that prediction varies a lot across users and thus, to eliminate bias introduced by specific users, we evaluate our model in a leave-one-out cross-validation fashion
using the following metrics:
\\
\noindent\textbf{Success/Failure accuracy ---} metric for monitor task. It shows how well the model discriminates a successful pouring sequence from a failed one. It directly indicates the performance of our main task.
\\
\noindent\textbf{Classification accuracy ---} metric for initial object state classification. It shows how well the model recognizes what kinds of container and amount of liquid in the containers in a pouring sequence.
\\
\noindent\textbf{Regression error ---} metric for trajectory forecasting. It is the error between 6-dimensional 3D trajectories recorded by HTC Vive and predicted 3D trajectories. Note that due to distinct properties of position and rotation error, the two errors are calculated separately. 

\gina{
\subsection{Setting Variants}\label{ssec:settingvariants}
To study the effectiveness of each independent component in our network, we evaluate different settings described below in the following experiments.  
\\
\noindent \textbf{Vanilla RNN}: Our fusion RNN without auxiliary tasks. The model is a LSTM encoder (see \autoref{ssec:LSTM}) followed by fully-connected layers. \ginarep{The LSTM encoder encodes the sequence of the image feature and sensor data. The fully-connected layers then perform success/failure classification based on the encoded features.}{The fully-connected layers  perform success/failure classification based on the encoded features.}  
\\
\noindent \textbf{RNN w/ IOSC}: Our fusion RNN with an auxiliary task, initial object state classification (IOSC). The details of IOSC are described in \autoref{ssec:isc}.  
\\
\noindent \textbf{RNN w/ TF}: Our fusion RNN with an auxiliary task, trajectory forecasting (TF). The details of TF are described in \autoref{ssec:tf}.
\\
\noindent \textbf{Ours w/o adv.}: Our fusion RNN with two proposed auxiliary tasks, initial object state classification and trajectory forecasting. In this setting, we treat one-step trajectory forecasting as a regression task (see \autoref{eq:regloss}).
\\
\noindent \textbf{Ours}: Our fusion RNN with two proposed auxiliary tasks, initial object state classification and trajectory forecasting. In this setting, we introduce the adversarial training loss (see \autoref{eq:adversarial}) to generate more diverse trajectories.
}

\begin{figure}[t!]
\centering
\resizebox{0.78\linewidth}{!}{
\centering
\begin{minipage}[c]{0.45\textwidth}
\centering
	\subfloat[\small Trajectory visualization]{\includegraphics[width=\textwidth]{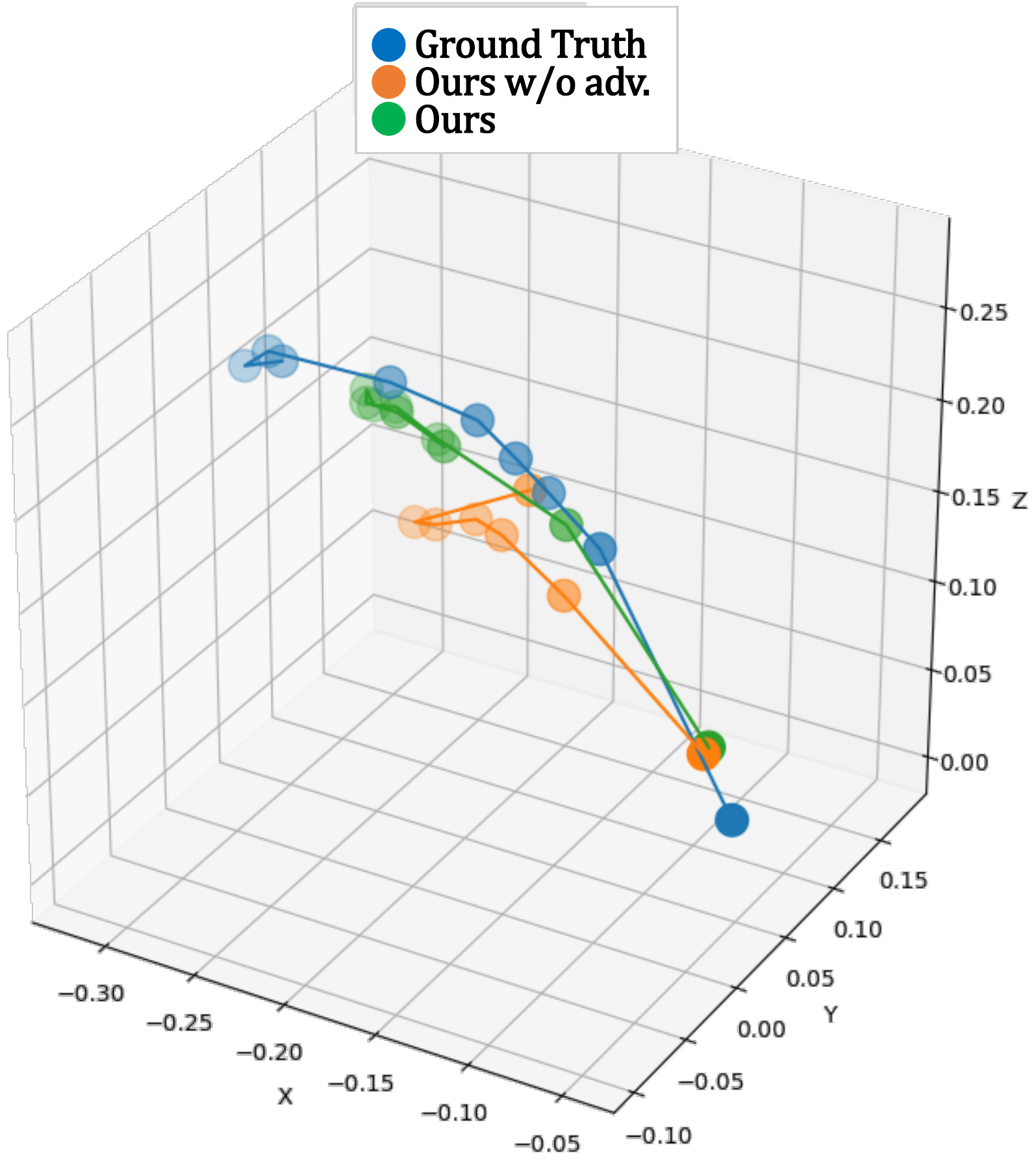}}
\end{minipage}
\begin{minipage}[c]{0.1\textwidth}
\end{minipage}
\begin{minipage}[c]{0.5\textwidth}
\centering
    \subfloat[\small Translation error]{\includegraphics[width=\textwidth]{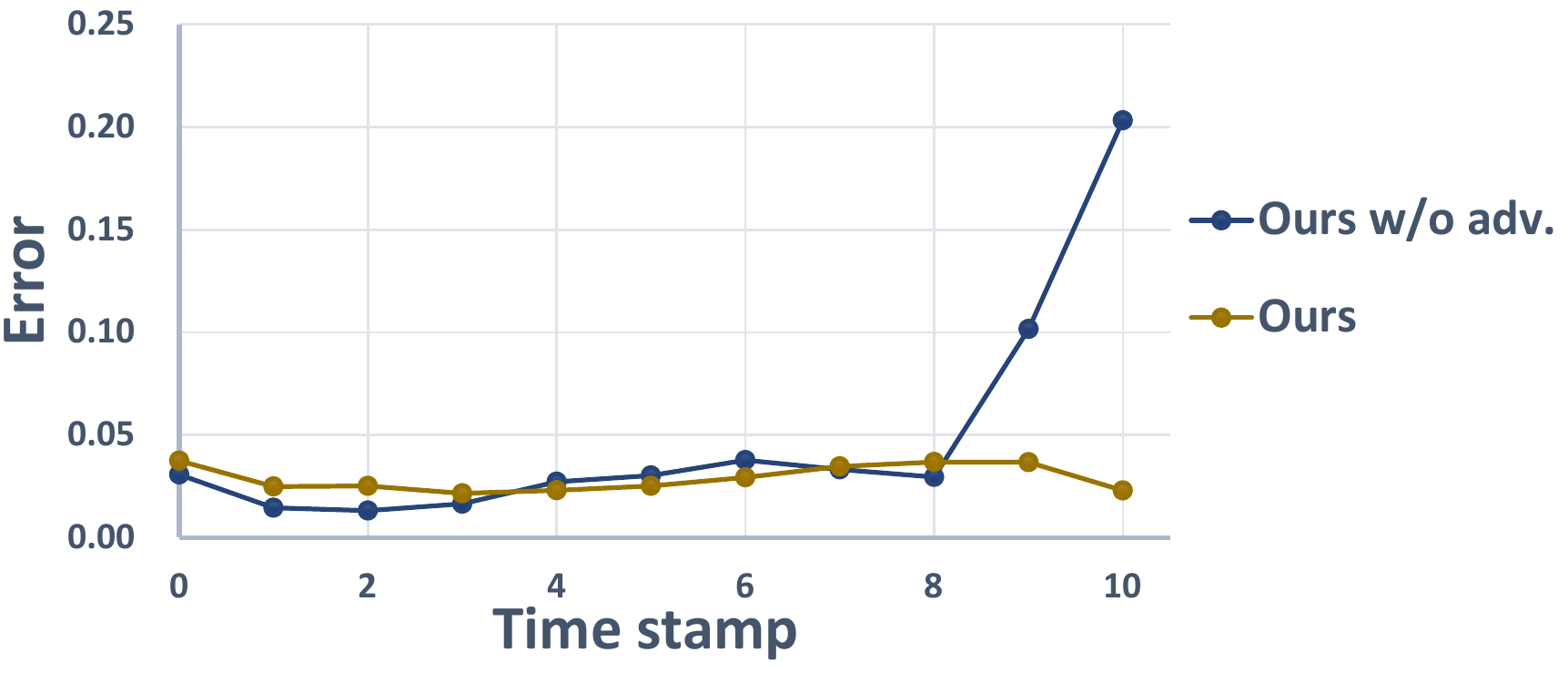}} \\
    \subfloat[\small Rotation error~~~]{\includegraphics[width=\textwidth]{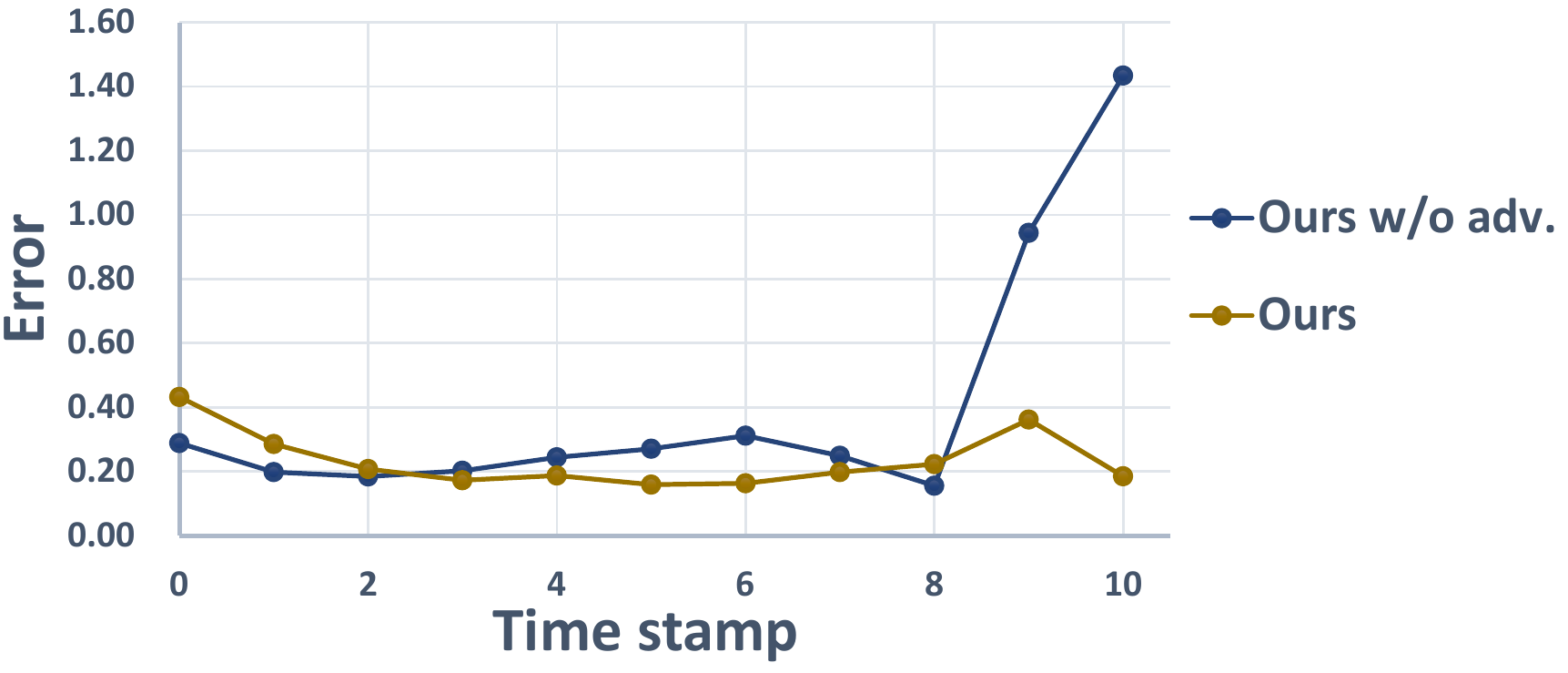}}
\end{minipage}
}
\caption{\textbf{Trajectory forecasting comparison} between ``$Ours~w/o~adv.$'' and ``$Ours$''. (a) Ground truth, ``$Ours~w/o~adv.$'' and ``$Ours$'' are shown in blue, orange and green, respectively. Time is visualized as color intensity goes from dark to light. Apparently, ``$Ours~w/o~adv.$'' failed to forecast the trajectory at a later stage of liquid pouring, while ``$Ours$'' can still follow the trend. (b)(c) ``$Ours$'' and ``$Ours~w/o~adv.$" have comparable errors at early steps, but the former one performs better in later steps}
\label{fig:tf}

\includegraphics[width=0.7\textwidth]{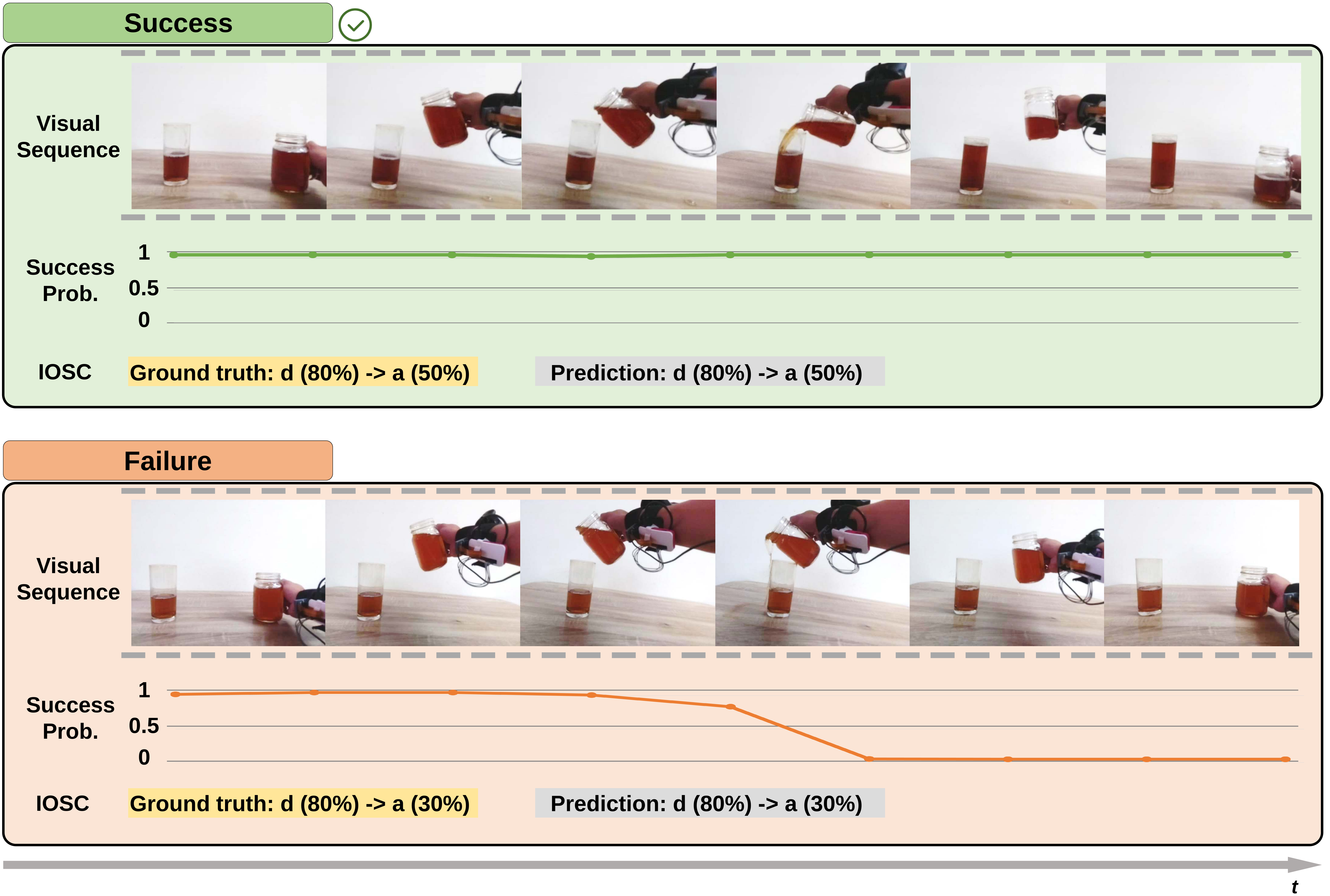}
\caption{\textbf{Monitoring along time.}
The prediction correctly follows the visual cues 
}
\label{fig:crossUserMonitoring}
\end{figure}

\gina{
\subsection{Monitoring Liquid Pouring}
We consider 3 scenarios to test our method's generalization ability. Firstly, we assume that our model is used to monitor a specific group of users with a specific set of containers. Then, in a more challenging scenario, we assume the model need to monitor unseen containers as well. Finally, we consider that the model needs to monitor unseen users. More details \ginarep{of these scenarios}{}are described below.
\\
\noindent \textbf{Cross Trial Experiment.}  
This experiment is the most simple case. Models are trained and tested on data of the same group of users with the same container set, but training data and testing data are collected from different trials of pouring.
In this easiest scenario, success/failure classification poses minor challenge here and is well solved. From \autoref{Tab:crossTrialResults}, we can see that our method generates better performance on monitoring than the baseline method (i.e. $vanilla~RNN$), which lacks two auxiliary tasks.
\\
\noindent \textbf{Cross Container Experiment.}  
This is a common scenario that may occur in the real use case. When using different containers to pour liquid, the whole pouring sequences may be very different. For instance, there are huge changes in the appearance and the pouring trajectories between the case of the teapot and the bottle.
We run leave-one-out cross-validation on the 4 different source containers to test whether our model can generalize to unseen containers. The initial states are only related to the liquid amount in the source ($10\%$, $50\%$, $80\%$) and target container ($0\%$, $30\%$, $50\%$), so we have 9 initial states (rather than 36 states) in total.
The results in \autoref{Tab:crossContainerResults} show that our method achieves better performance on monitoring than the baseline method, since it successfully catches the change of states and the hand dynamics during the pouring sequence.
\\
\noindent \textbf{Cross User Experiment.}
This is the most challenging scenario, since different demonstrators may have very different pouring \jc{styles}.
\gina{Considering a specific set of containers,} \ginarep{Models}{models} are trained on data of 4 different users and tested on 1 user other than the 4 users in training set. The main difference among cross-user data is the variance in pouring \jc{styles}. To be more precise, this experiment examines generalization ability in IMU sensor data sequences.
By looking at success/failure accuracy shown in \autoref{Tab:crossUserResults}, we can find that both auxiliary tasks, initial-state classification and trajectory forecasting, brings considerable improvement in monitoring object manipulation.
From \autoref{fig:crossUserMonitoring}, we can observe that our model's prediction correctly follows the visual cues.
Initial object state classification helps the model know what the source container and the target container are, and the amount of liquid in both containers. Trajectory forecasting helps the model learn local dynamics of pouring sequences. Remarkably, by comparing our method and $Ours~w/o~adv.$, we can find that adversarial training introduced in our method significantly boosts initial state classification and slightly improves trajectory forecasting. From the results, we infer that there is implicitly-shared knowledge between the two auxiliary tasks and a more robust trajectory forecasting may enhance initial state classification. Adversarial training does help regarding obtaining a better understanding of pouring behaviors and increase the performance of our model in monitoring task. 
}

\begin{table}[t!]
\centering
\caption{The results of cross trial experiments 
}
\label{Tab:crossTrialResults}
\resizebox{0.8\linewidth}{!}{
\begin{tabular}{|c||c|c|c|c|}
\hline
 & succ./fail. acc. & classification acc. & position error & rotation error\\ \hline
$Vanilla~RNN$ & $99.65$ \% & $N/A$ & $N/A$ & $N/A$\\ \hline
$Ours~w/o~adv.$ & $100$ \% & $96.50$ \% & $0.020~m$ & $7.58^{\circ}$ \\ \hline
$Ours$ & $100$ \% & $96.07$ \% & $0.020~m$ & $\textbf{6.80}^{\circ}$ \\
\hline
\end{tabular}
}

\begin{center}
\caption{The results of cross container experiments 
}
\label{Tab:crossContainerResults}
\resizebox{0.8\linewidth}{!}{
\begin{tabular}{|c||c|c|c|c|}
\hline
 & succ./fail. acc. & classification acc. & position error & rotation error\\ \hline
$Vanilla~RNN$ & $89.16$ \% & $N/A$ & $N/A$ & $N/A$\\ \hline
$Ours~w/o~adv.$ & $96.45$ \% & $63.92$ \% & $0.040~m$ & $11.11^{\circ}$ \\ \hline
$Ours$ & $\textbf{97.11}$ \% & $\textbf{67.69}$ \% & $\textbf{0.038}~m$ & $11.30^{\circ}$ \\ \hline
\end{tabular}
}
\end{center}

\begin{center}
\caption{The results of cross user experiments 
}
\label{Tab:crossUserResults}
\resizebox{0.8\linewidth}{!}{
\begin{tabular}{|c||c|c|c|c|}
\hline
 & succ./fail. acc. & classification acc. & position error & rotation error\\ \hline
$Vanilla~RNN$ & $81.95$ \% & $N/A$ & $N/A$ & $N/A$ \\ \hline
$RNN~w/~IOSC$ & $89.25$ \% & $68.51$ \% & $N/A$ & $N/A$ \\ \hline
$RNN~w/~TF$ & $90.82$ \% & $N/A$ & $0.033~m$ & $14.15^{\circ}$ \\ \hline
$Ours~w/o~adv.$ & $92.97$ \% & $64.15$ \% & $0.033~m$ & $14.20^{\circ}$ \\ \hline
$Ours$ & $\textbf{93.25}$ \% & $\textbf{75.69}$ \% & $0.033~m$ & \textbf{14.06}$^{\circ}$ \\
\hline
\end{tabular}
}
\end{center}

\centering
\caption{Ablation study on LSTM architecture 
}
\label{Tab:lstmAblation}
\resizebox{0.8\linewidth}{!}{
\begin{tabular}{|c||c|c|c|c|}
\hline
LSTM architecture & succ./fail. acc. & classification acc. & position error & rotation error\\ \hline
$2\text{-}layer$  & $87.06$ \% & $58.92$ \% & $0.033
~m$ & $14.72^{\circ}$ \\ \hline
$hierachical$ & $\textbf{93.25}$ \% & $\textbf{75.69}$ \% & $0.033~m$ & $\textbf{14.06}^{\circ}$ \\
\hline
\end{tabular}
}
\end{table}

\gina{
\subsection{Discussion}\label{ssec:discusion}
In this section, we further discuss each component in our network and the future feasibilities. Firstly, we do ablation study on LSTM architecture under the cross-user scenario, comparing the hierarchical LSTM (see \autoref{ssec:LSTM}) to a 2-layer LSTM. The latter one is an early fusion method that data from different modalities is directly concatenated together and fed into the 2-layer LSTM. The results in \autoref{Tab:lstmAblation} show that the hierarchical LSTM with late fusion outperforms the naive 2-layer LSTM in all tasks and this may be due to the capability of the hierarchical LSTM to handle scale difference and imbalanced dimension among multimodal inputs.

Secondly, we study the effect of the adversarial loss to the whole network. Recall that we introduce adversarial loss since there are multiple feasible trajectories for each data sample. However, these errors assume that there is only one truth position and rotation of each testing sample.
As mentioned above, our model learns a more general concept and will predict trajectory based on common knowledge considering pouring, whereas prediction of ``$Ours~w/o~adv.$" heavily relies on knowledge of seen trajectories and will drastically fail if testing pouring sequences have little in common with training data. This can be observed in \autoref{fig:tf}.a. Also, the adversarial loss will allow the model to generate more diverse trajectories, which means the model will observe more diverse hidden states in later steps. The trajectory forecasting errors in \autoref{fig:tf}.b and \ref{fig:tf}.c show that ``$Ours$'' and ``$Ours~w/o~adv.$" have comparable errors at early steps, but the former one perform better in later steps. 

Our experiments show that introducing auxiliary tasks is beneficial for understanding the subtle liquid pouring task. By implicitly modeling the environmental states and hand dynamics, we improve liquid pouring monitoring significantly. We believe the general idea applies to other subtle manipulating tasks like opening doors\gina{, driving nails} and cutting bread. Intuitively speaking, opening doors also involves mapping visual (e.g., what types of doors) and non-visual (e.g., hand motion) observations into environmental states to facilitate monitoring whether the door is opened. 
\ginarep
{In computer vision, we are early to propose a system to jointly collect visual and non-visual observations at a scale (1800 trials). In the future, we plan to utilize more sophisticated non-visual sensors on other subtle manipulation tasks. For instance, IMU on the wrist might be too limited for tasks involving finger level motion.}{
Monitoring different tasks may need different auxiliary tasks to make use of rich sensories in order to learn both visual and non-visual signals.}
}

\section{Conclusion}
\label{sec:con}
\gina{
In this work, we aim at learning to monitor whether liquid pouring is successful (e.g., not spilling) or not using synchronized visual and IMU signals. We propose a novel method containing two auxiliary tasks during training: inferring (1) the initial state of containers and (2) forecasting the one-step future 3D trajectory of the hand with an adversarial training procedure.
These tasks encourage our method to learn representation sensitive to container states and how objects are manipulated in 3D.
On our newly collected liquid pouring dataset, our method achieves $\sim 8\%$ and $\sim 11\%$ better monitoring accuracy than the baseline method without auxiliary tasks on unseen containers and unseen users respectively.
}

\noindent \textbf{Acknowledgements}  
We thank Stanford University for collaboration. We also thank MOST 107-2634-F-007-007, Panasonic and MediaTeK for their support.


\clearpage


\end{document}